\documentclass[letterpaper]{article} 
\usepackage{aaai24}  
\usepackage{times}  
\usepackage{helvet}  
\usepackage{courier}  
\usepackage[hyphens]{url}  
\usepackage{graphicx} 
\usepackage{natbib}  
\usepackage{caption} 
\usepackage{amsmath, amsfonts}
\usepackage{xcolor}         
\usepackage{natbib}
\usepackage{graphicx}
\usepackage{amsmath}
\usepackage{tabularx}
\usepackage{algorithmic}
\usepackage{textcomp}
\usepackage{xcolor}
\usepackage{float}
\usepackage{multirow}
\usepackage{paralist}
\usepackage{comment}
\usepackage{ragged2e}
\usepackage{algorithm}
\usepackage{algorithmic}

\usepackage{newfloat}
\usepackage{listings}
\DeclareCaptionStyle{ruled}{labelfont=normalfont,labelsep=colon,strut=off} 
\floatstyle{ruled}
\newfloat{listing}{tb}{lst}{}
\floatname{listing}{Listing}
\title{SC-Phi2: A Fine-tuned Small Language Model for StarCraft II Macromanagement Tasks}

\title{SC-Phi2: A Fine-tuned Small Language Model for StarCraft II Macromanagement Tasks}

\author {
    Muhammad Junaid Khan,
    Gita Sukthankar
}
\affiliations {
    Department of Computer Science, University of Central Florida, USA\\
    mu718889@ucf.edu, gita.sukthankar.ucf.edu
}

\usepackage{bibentry}

\begin{document}

\maketitle

\begin{abstract}
This paper introduces SC-Phi2, a fine-tuned StarCraft II small language model for macromanagement tasks. Small language models, like Phi2, Gemma, and DistilBERT, are streamlined versions of large language models (LLMs) with fewer parameters that require less power and memory to run. To teach Microsoft's Phi2 model about StarCraft, we create a new SC2 text dataset with information about StarCraft races, roles, and actions and use it to fine-tune Phi-2 with self-supervised learning. We pair this language model with a Vision Transformer (ViT) from the pre-trained BLIP-2 (Bootstrapping Language Image Pre-training) model, fine-tuning it on the MSC replay dataset.  This enables us to construct dynamic prompts that include visual game state information. Unlike the large models used in StarCraft LLMs such as GPT-3.5, Phi2 is trained primarily on textbook data and contains little inherent knowledge of StarCraft II beyond what is provided by our training process. By using LoRA (Low-rank Adaptation) and quantization, our model can be trained on a single GPU.  We demonstrate that our model performs well at micromanagement tasks such as build order and global state prediction with a small number of parameters.  

\end{abstract}

\section{Introduction}

\citeauthor{gallotta2024largelanguagemodelsgames} (\citeyear{gallotta2024largelanguagemodelsgames}) charted a course for the application of LLMs to games, examining their performance as players, non-player characters, commentators, game masters, and designers.  Despite their successes, LLMs often experience continuity problems, due to constraints on context size. However their ability to perform commonsense reasoning makes them a natural fit for open-world games like Minecraft, as demonstrated by recent studies~\cite{voyager, survey_llm, mc_ghost, minedreamer, open_world_long_horizon}. 

\citeauthor{Kambhampati_2024}  (\citeyear{Kambhampati_2024}) asserts that LLMs have very limited reasoning ability and fail at basic planning tasks if small perturbations are made. This makes strategic games, like the highly demanding real-time strategy  game StarCraft II (SC2), a difficult area for them. SC2 requires players to excel at both high-level \textit{macromanagement} tasks like production strategies along with \textit{micromanagement} (unit-level tactics). The most successful AI systems for SC2 are AlphaStar \cite{alpha} and ROA-Star \cite{roa_star}, both of which have been trained using imitation learning and reinforcement learning techniques. These systems can play full games and achieve Grandmaster-level performance but require substantial computational resources, needing at least 64 Nvidia V100 GPUs for training.

Using human evaluators, \citeauthor{textsc2} (\citeyear{textsc2}) tested different LLMs on their understanding of StarCraft concepts such as game rules, race mechanics, build orders, and strategies; they showed that GPT-4 and 3.5 can accurately answer detailed questions about 
SC2 play.  However these models are extremely large; for instance, GPT-4 has 1 trillion parameters. This paper proposes the usage of a small language model, Phi-2~\cite{phi2} for StarCraft macromanagement tasks.  Small language models (SLMs) are streamlined cousins of large language models that are trained on more curated datasets. They offer lower processing latency, making them well suited for chatbots, mobile devices, and real-time applications such as games. Although Phi-2 does not inherently know much about StarCraft II, it excels at commonsense reasoning tasks and only has 2.8 billion parameters.  To teach Phi-2 about SC2, we created a new text dataset with information about StarCraft races, roles, and strategies to fine-tune the model using supervised learning.  This paper shows that it is feasible to use a significantly smaller autoregressive language model for StarCraft II macromanagement than has been demonstrated in previous work~\cite{textsc2,swarmbrain}.




Our proposed architecture integrates the Microsoft Phi-2 language model \cite{phi2} with the Vision Transformer (ViT) from BLIP-2 \cite{blip2}. The training process is conducted in two stages: i) Stage 1 focuses on fine-tuning the Phi-2 model on our SC2 Text Dataset to provide the language model with detailed knowledge of SC2 gameplay; and ii) Stage 2 further fine-tunes the model for various match-ups using separate Parameter Efficient Fine-Tuning (PEFT) adapters for each match-up, specifically targeting build order prediction and game state prediction.

\begin{figure*}[!ht]
    \centering
    \includegraphics[width=0.8\linewidth]{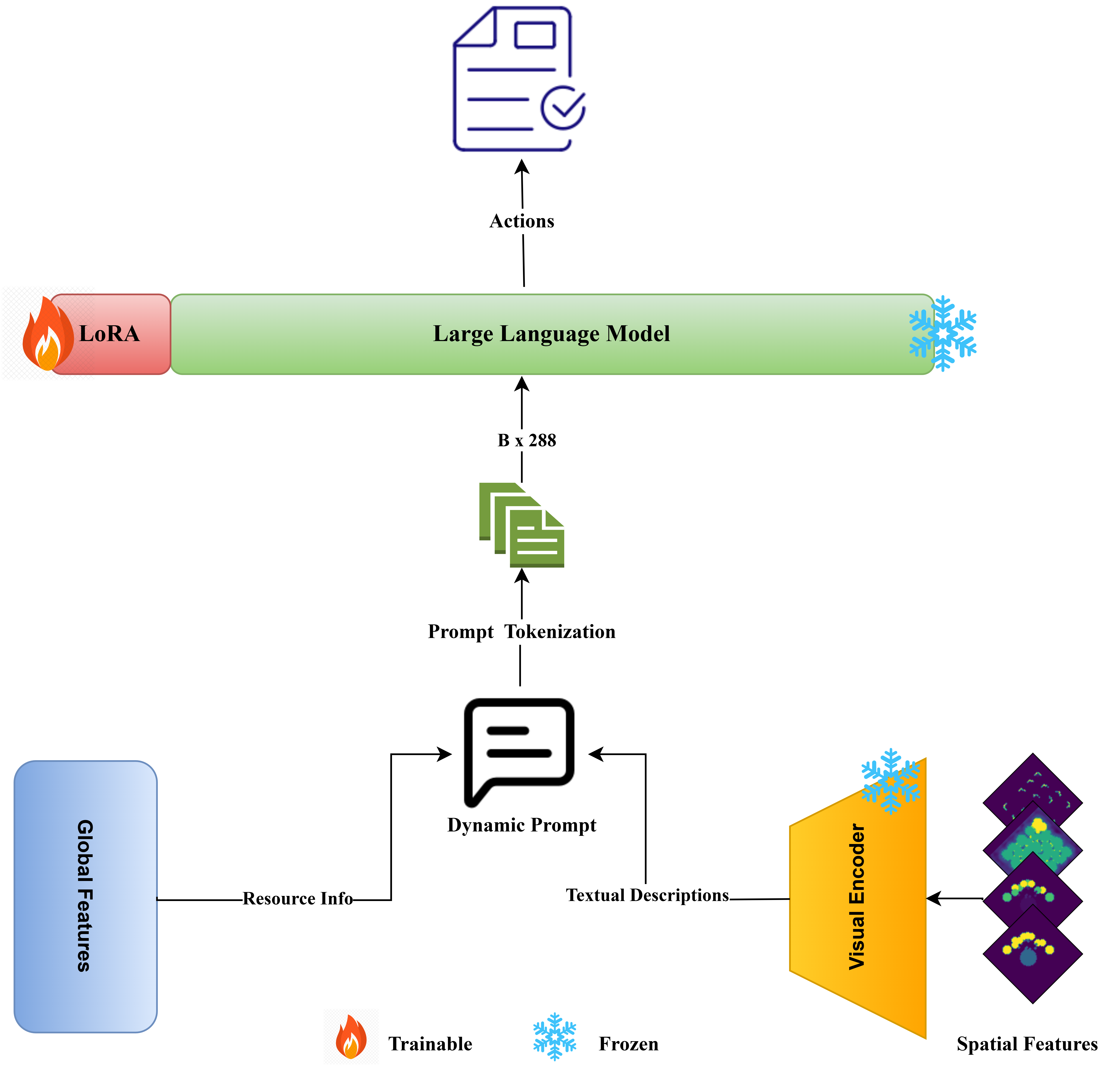}
    \caption{SC-Phi2 Model. \textit{Spatial features} represent screen and mini-map features while \textit{global features} represent supplies and scores. During the training, we construct a dynamic prompt from both the global features and the textual descriptions generated by the pre-trained Vision Encoder, ViT, from the BLIP-2 vision-language model. Here, we use fine-tuned Phi-2 from stage 1 of fine-tuning, again fine-tuning about $4\%$ of parameters using the LoRA approach.}
    \label{fig:scgpt}
\end{figure*}

During Stage 1, we fine-tune only the Phi-2 language model. In Stage 2, we incorporate both the Phi-2 model and the textual descriptions of visual features extracted by the ViT model. We design dynamic prompts that include gameplay details such as game stage, resources, army buildings, and food, combined with these textual descriptions. Based on these dynamic prompts, our model predicts the next actions (i.e., the build order) and the game outcome (i.e., win or loss).

Our key contributions are: i) the development of an SC2 text dataset for instructional fine-tuning of the language model; ii) the introduction of an SLM-based multimodal approach for macromanagement prediction; and iii) the achievement of these results on a single GPU using PEFT and quantization techniques.

\section{Related Work}

\subsection{LLMs in StarCraft}
StarCraft game replays contain a large amount of spatial data that is not easily accessible to LLMs.  To overcome this problem \citeauthor{textsc2} (\citeyear{textsc2}) introduced a text-based interface, TextStarCraft II that has specialized adapters for mapping observations to text and text to actions. Even with these adapters, it is infeasible for LLMs to operate at the frame rate speeds necessary for SC2 agents. To combat the speed problem, the authors experimented with multi-frame summarization techniques and action queues.  Using this text version of the game, \citeauthor{textsc2} (\citeyear{textsc2}) showed that GPT-3.5, which has 175 billion parameters, performs at a level comparable to a mid-range human player. The authors suggested that incorporating visual data could improve system performance; our proposed model, SC2-Phi2, uses a pre-trained vision transformer.

SwarmBrain~\cite{swarmbrain} is a more specialized agent that also uses GPT-3.5 to make strategic decisions for SC2~\cite{swarmbrain}. In SwarmBrain, the Overmind Intelligence Matrix focuses on high-level strategic decisions like resource allocation and base expansion, while the Swarm ReflexNet handles immediate tactical responses in battle.  In our work, we demonstrate that a significantly smaller model, with only 2.8 billion parameters, can be used for macromanagement tasks, with the right fine-tuning. 


\subsection{MSC Dataset}
The MSC dataset, introduced by \citeauthor{msc} (\citeyear{msc}), is based on the SC2LE \cite{sc2le} and comprises over 36,000 replays. It serves as a comprehensive resource for training and evaluating machine learning models for macromanagement tasks in StarCraft II (SC2). To ensure the quality and relevance of the replays, a rigorous preprocessing pipeline was implemented, ensuring that each replay meets the following criteria:
\begin{itemize}
    \item Each match within the replay contains at least 10,000 or more frames.
    \item Both the player and the opponent have at least 10 APM (actions per minute) rate.
    \item Both players have at least 1000 MMR (match-making ratio).
    \item Broken or incomplete replays are excluded.
\end{itemize}

Each replay includes global features such as resources collected, and detailed information about units and buildings, all normalized between 0 and 1. Additionally, each replay contains spatial features with a shape of $\mathbb{R}^{13 \times 64 \times 64}$. The final outcome of each match, whether a win or a loss, is also recorded and represented by 1 and 0, respectively.  This dataset has been used by several others to evaluate build order and win/loss prediction. This paper benchmarks the prediction capabilities of SC-Phi2 against the other two top performers~\cite{msc,junaid0}.

\section{Method}
Our proposed method operates in two distinct stages:

i) Stage 1: Primarily concentrates on fine-tuning the Microsoft Phi-2 model \cite{phi2} utilizing our proposed SC2 dataset. This initial stage is dedicated to optimizing model performance specifically for SC2-related tasks.

ii) Stage 2: Proceeds with additional fine-tuning of the Phi-2 model using the MSC dataset. Notably, this stage incorporates textual descriptions sourced from a pre-trained ViT encoder from the BLIP-2 model \cite{blip2}. The integration of ViT embeddings enriches the model's understanding of textual context from spatial features, enhancing its overall performance.


\subsection{SC2 Text Dataset}


While LLMs can generate general information about SC2, optimizing the Phi-2 model for SC2-specific tasks demands a nuanced approach. To achieve this, we create a specialized text dataset tailored precisely to model the complexity of SC2 gameplay. Our dataset was aggregated from several online SC2 resources~\cite{liquipedia, scwiki,scwikipedia} and covers the essential elements of SC2 gameplay. It encompasses extensive details on the Protoss, Terran, and Zerg races, capturing their unique characteristics, building specifications, and unit tactics. It also includes specific information on the strengths, weaknesses, and special abilities of each unit.

By providing a thorough understanding of these elements, the dataset facilitates strategic planning and decision-making, and can be used for instructional fine-tuning of the language model.
Also the dataset includes online sources discussing the roles and effectiveness of units against different opponents, offering invaluable insights for developing effective combat strategies. This includes an analysis of unit interactions and counter-strategies, helping to predict and adapt to enemy tactics.

Beyond race-specific details, our dataset incorporates rich information on common actions drawn from the PySC2 library \cite{sc2le}. This encompasses a wide range of actions and maneuvers crucial for effective gameplay, such as unit commands, building, training, and morphing actions, and resource management techniques. By including these practical in-game actions, the dataset is enriched with actionable knowledge that mirrors real gameplay scenarios.  Lastly, the dataset also contains some common build orders for each of these races, providing the model to learn effective build order strategies. Details of the dataset are provided in the appendix, and the dataset itself is available by request.

By compiling such a comprehensive repository of SC2-related information, we lay the groundwork for refining the Phi-2 model. This enables us to enhance its performance on SC2-specific tasks, ensuring that the model can accurately interpret and respond to a diverse range of in-game situations. The rich, detailed dataset not only supports the development of a more robust macromanagement system but also provides a valuable resource for ongoing SC2 research. 

\begin{figure}
    \centering
    \includegraphics[width=1\linewidth]{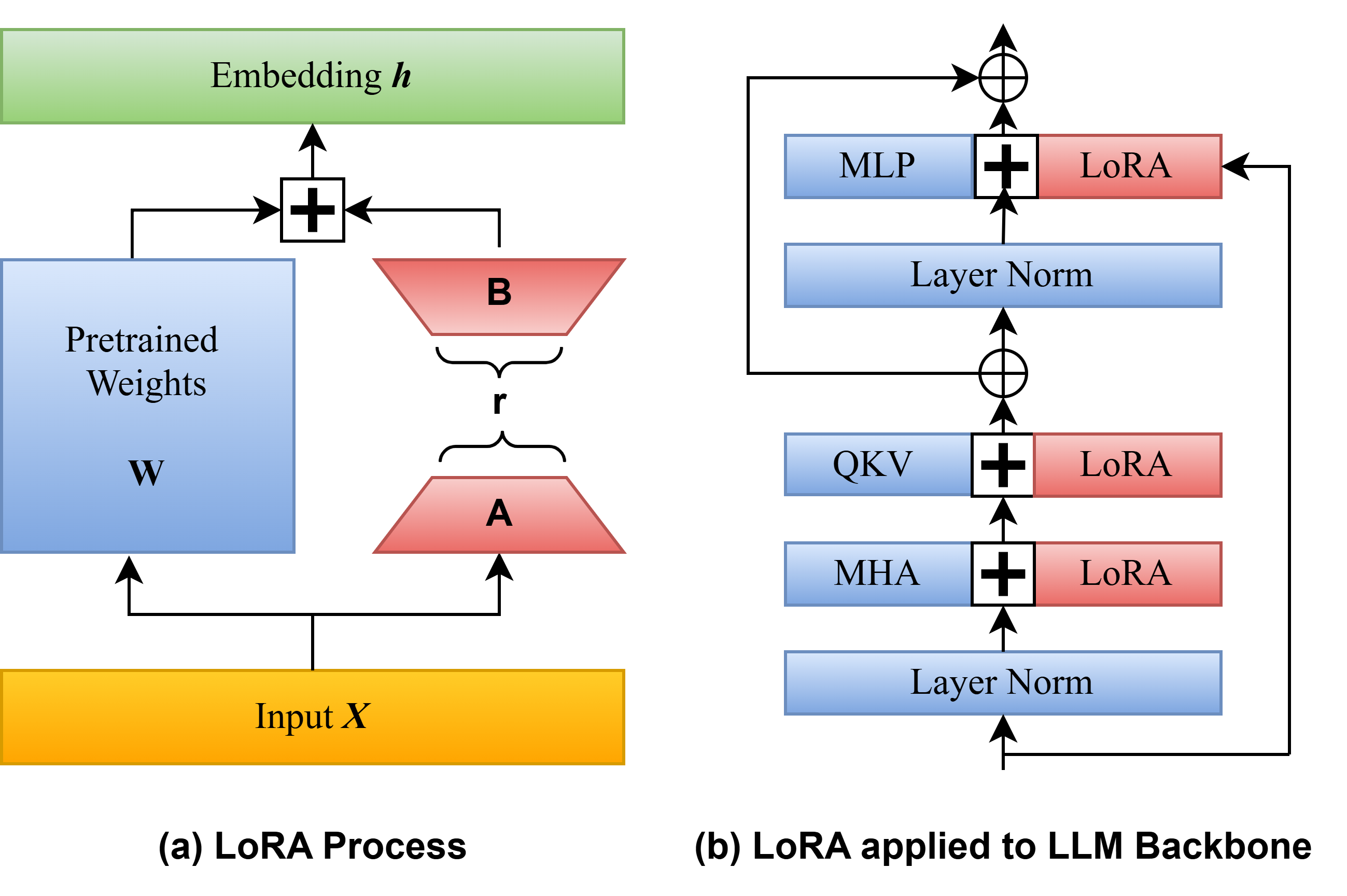}
    \caption{LoRA Adaptation for Language Backbone. (a) shows the general LoRA process. (b) LoRA applied to specific layers in our approach. In the diagram, the red blocks represent the weights updated during training, while the blue blocks denote the frozen weights. A and B are low rank matrices and \textbf{r} is a LoRA hyper-parameter.}
    \label{fig:lora}
\end{figure}

\begin{figure*} [!htp]
    \centering
    \includegraphics[width=1\linewidth]{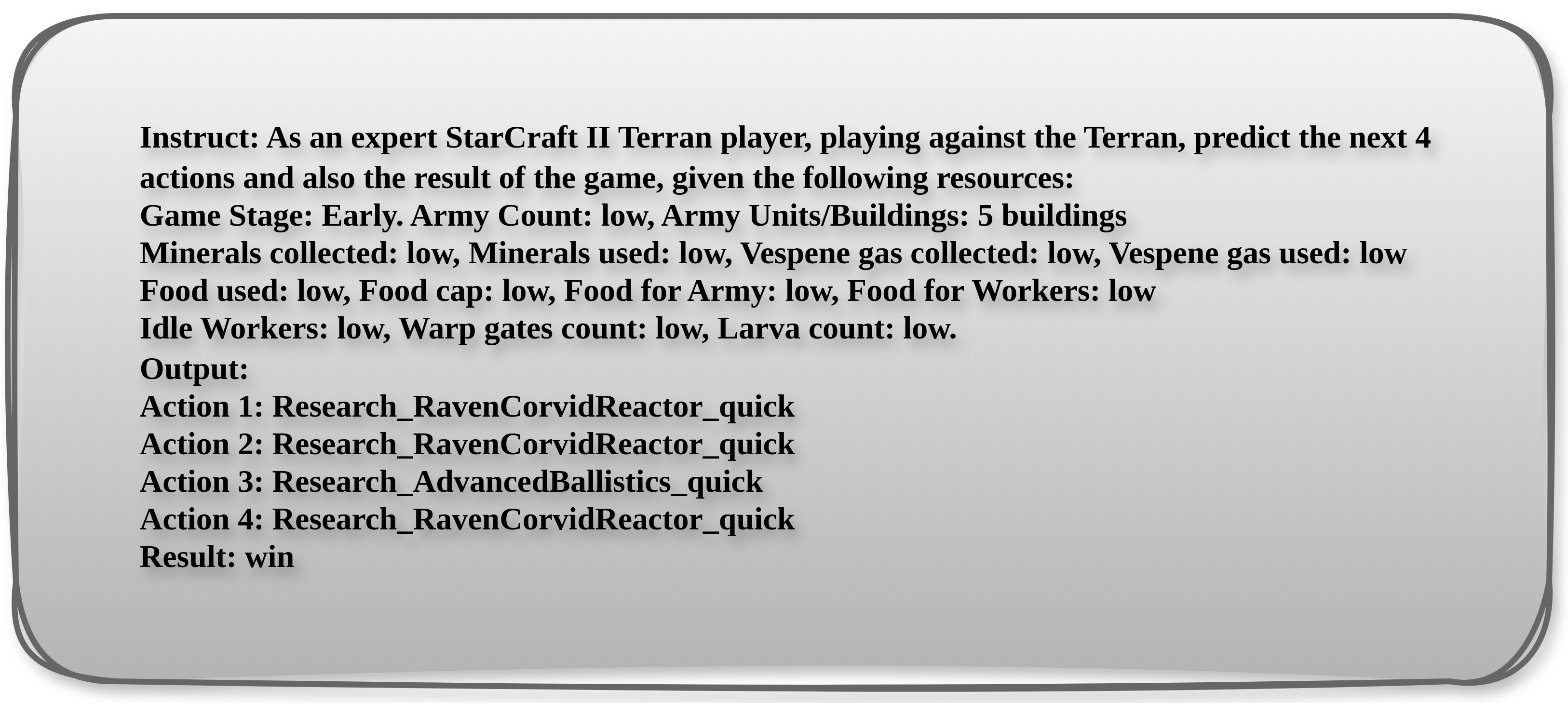}
    \caption{Prompt used during the stage-2 fine-tuning. Numerical values have been changed to categorical values during training. For example, the value of feature \textbf{Game Stage} is \textbf{Early}\textbf, and {Army Count} is \textbf{low} in the prompt. Similarly all other values have been changed. }
    \label{fig:s2_prompt}
\end{figure*}

\subsection{Stage-1 Fine-tuning the SLM}
In Stage 1, we focus on the Self-Supervised Fine-Tuning of the Phi-2 model using our SC2 Text Dataset. To facilitate this fine-tuning process, we leverage the SFTTrainer module provided by the Hugging Face Transformers library \cite{huggingfaces}.

Despite its relatively smaller scale, with 2.8 billion parameters compared to other language models, Phi-2 has consistently demonstrated superior performance over larger counterparts across numerous language tasks. This underscores the efficiency and effectiveness of Phi-2 as a foundation for our model, showcasing its efficacy in handling complex linguistic tasks with remarkable accuracy. 

During this stage, we specifically fine-tune the attention layers and feed-forward fully connected layers of the transformer layers within the Phi-2 model using both Low Rank Adaptation (LoRA) and the Quantized Low Rank Adaptation (QLoRA) approach \cite{lora, qlora}. This fine-tuning process is illustrated in Figure \ref{fig:lora}. For Stage 1, we specify the LoRA parameters as \textbf{alpha = 128} and \textbf{r = 64}. 

Regarding quantization, we load the model and optimizer in 8-bit mode, conducting fine-tuning over 160 epochs. Additionally, we include a comparative analysis of various configurations across different sets of hyperparameters in the appendix (Table \ref{table:app_stg1} and \ref{table:t_params}). Utilizing both the LoRA and QLoRA approaches enable us to fine-tune our entire model efficiently on a single GPU.

\subsection{Stage-2 Fine-tuning}

Stage 2 of the fine-tuning process starts with merging the QLoRA-based adapter and its weights trained during stage 1 into the main Phi-2 model. This integration enhances the efficiency of both fine-tuning and inference operations during Stage 2. We keep most of the hyperparameters from stage 1 unchanged except the text prompt, the length of tokens, which is set to 288 tokens, and the batch size, which is set to four. 

Leveraging the MSC dataset, we fine-tune the model once again, following a self-supervised approach. Furthermore, for this stage, in addition to Phi-2, we incorporate a Vision Transformer (ViT)-based vision encoder sourced from the BLIP-2 model. The subsequent sections provide detailed insights into the functionality and implementation of each component.  

\subsubsection{Visual Backbone}
Vision Transformers have emerged as a pivotal tool in vision-language tasks, owing to their exceptional effectiveness and versatility \cite{tinygptv, blip2, clip}. To complement the language backbone, we employ a pre-trained Vision Transformer (ViT) \cite{vit} sourced from the vision-language model BLIP-2 \cite{blip2} as our visual backbone. 

This particular version of ViT model is pre-trained on a diverse array of vision-language tasks, equipping it with the capability to provide textual descriptions for input images/frames. By incorporating this ViT model into our architecture, we aim to enhance our model's ability to interpret visual cues and seamlessly integrate them into our Dynamic Prompt generation setup. 

During the fine-tuning process, we input the map screen features into the visual backbone to extract their textual descriptions. For instance, circles on the map screen represent buildings, and these textual descriptions offer valuable insights into the current state of army buildings. These insights are then incorporated into our dynamic prompt generation step, creating a prompt for fine-tuning the model. This integration is shown in Figure \ref{fig:s2_prompt}.

\subsubsection{Global Features for Prompt Generation}

Integral to our approach is the incorporation of global features extracted from the MSC dataset into our Dynamic Prompt Generation component. These features encompass critical aspects of the game such as food information, army building progress, and resource collection rates. These features also serve as invaluable indicators, offering essential insights into the evolving dynamics of the game. By assimilating this multifaceted information, our model gains a comprehensive understanding of the complex gameplay dynamics inherent in SC2, empowering it to make informed decisions and predictions with high accuracy.

It is important to note that these global features in the MSC dataset are numerical values, with most of them normalized between 0 and 1. To enhance the model's representation and context, we convert these numerical values into categorical values. Specifically, the category \textbf{low} corresponds to numerical values between 0 and 0.2, the category \textbf{medium} represents values between 0.21 and 0.7, and the category \textbf{high} is assigned to values greater than 0.7. Similarly, the game stage is categorized into four distinct phases: i) \textbf{Early}: representing the initial stage of the game; ii) \textbf{Mid}: indicating the progress of the game between 25\% and 60\%; iii) \textbf{Late}: covering the period between 60\% and 90\%; and iv) \textbf{End}: denoting the final phase, over 90\% of the game.

Additionally, we redefine the rewards, converting a reward of 0 to \textbf{loss} and a reward of 1 to \textbf{win}. The actions are also transformed from action IDs to their corresponding full action names. For example, the action ID \textbf{75} is mapped to the action \textbf{Build\_Reactor\_Factory\_quick} according to the PySC2 library.

\begin{table} [!ht]
\centering
\renewcommand{\arraystretch}{1.5}
\begin{tabularx}{0.45\textwidth} { 
  >{\raggedright\arraybackslash}X 
  | >{\centering\arraybackslash}X
  | >{\centering\arraybackslash}X}
 \textbf{Match-up} & \textbf{No. of Replays} & \textbf{No. of Replays Used} \\
 \hline
 \hline
  Terran vs Terran (TvT) & 4897 & 1000 \\ 
  \hline
  Protoss vs Protoss (PvP) & 4334 & 1000\\
  \hline  
  Zerg vs Zerg (ZvZ) & 2989 & 1000
\end{tabularx}
\caption{Number of MSC replays for each racial match-up and the no. of replays used for 
fine-tuning.} 
\label{table:replays}
\end{table}

\subsubsection{Dynamic Prompt Generation}
In our methodology, we employ dynamic prompts generated during the training process. These prompts are crafted utilizing global features, including vital information such as food availability, army status, mineral and vespene gas reserves, and textual information of spatial features generated from the vision encoder. 

As the game progresses, these feature values dynamically evolve. To effectively adapt to these fluctuations, we continuously update the prompts in real-time, ensuring that the language model remains informed about the prevailing game circumstances. By providing this contextualized feedback, our model gains deeper insights into the evolving dynamics of the game, enhancing its ability to make informed decisions and predictions.  The training prompt is shown in Figure \ref{fig:s2_prompt}.

\subsubsection{Prompt Strategy} 
In our approach, we utilize a simple prompt for both training and evaluation, in contrast to Chain of Thought~\cite{cot} and other advanced prompt engineering techniques often employed in related works. While these sophisticated approaches are highly effective, especially with large models like GPT-4, they may not be as suitable for smaller models like Phi-2. Given that Phi-2 is significantly smaller and not trained at the same scale as GPT-4, a simpler prompt proves to be more effective for fine-tuning in our context.

Additionally, our method offers a comprehensive mechanism for fine-tuning a multimodal model on a single GPU. This approach not only ensures efficient training but also allows for further reduction of computational load during inference, enhancing the overall feasibility and performance of the model. 

\subsubsection{Final Fine-tuning of SLM}

After generating dynamic prompts, our SLM undergoes another round of fine-tuning, this time utilizing these dynamic prompts as inputs. This fine-tuning process follows the same strategy as Stage 1, employing the QLoRA approach with the optimal set of hyperparameters identified in Stage 1. Once the fine-tuning is complete, we merge these adapters back into the main Phi-2 model, enhancing its inference capacity. The architecture of our model is presented in Figure \ref{fig:scgpt}. 

\subsection{Training}
Fine-tuning LLMs can be challenging due to the problem of encountering 'NaN' and 'Inf' values during the backward pass. To ensure the consistent behavior of both the training and the evaluation phases, we set seed values for both the Numpy and the Pytorch libraries. In addition, enabling anomaly detection within PyTorch's autograd engine helps in rapidly identifying and addressing any computational anomalies that may arise.

We adopt a strategy where only a small fraction (approximately 4\%) of the parameters of the pretrained Phi-2 model are fine-tuned in each stage. This fine-tuning process is executed utilizing both the LoRA and the QLoRA approaches, which efficiently updates the model's parameters to adapt to the specifics of the SC2 domain. This process is illustrated in Figure \ref{fig:lora}. Meanwhile, we freeze the visual encoder and token generation components throughout the training process. 

The MSC dataset provides over 36000 pre-processed game replays for training the model. However, we only utilize a small subset of the replays to fine-tune our model. These details have been listed in Table \ref{table:replays}. 

\subsubsection{LoRA and QLoRA Adaptation for Language Backbone}
Fine-tuning LLMs can pose significant challenges and often demands substantial computational resources. Moreover, fine-tuning LLMs with multiple modalities can exacerbate issues such as gradient vanishing during training, as highlighted by \citeauthor{tinygptv} (\citeyear{tinygptv}). To mitigate such challenges, we employ the LoRA process along with its variant QLoRA in our language backbone.

LoRA is a mathematical technique aimed at reducing the number of trainable parameters in a model. Unlike traditional fine-tuning methods that update the entire model, LoRA adaptation selectively updates only a small subset of the model's parameters~\cite{lora}. For any specific layer, the weights are updated using LoRA process as:
\[
    \mathcal{W}_{0} + \Delta\mathcal{W} = \mathcal{W}_{0} + AB
\]

In this equation, $\mathcal{W}_{0}$ represents the pretrained weights of the large model, while $\Delta\mathcal{W}$ represents the updated weights obtained through low-rank matrices $A$ and $B$. Here, $\mathcal{W}_{0} \in \mathcal{R}^{d \times k}$, $A \in \mathcal{R}^{r \times k}$, and $B \in \mathcal{R}^{d \times r}$, and $rank \quad r << min(d, k)$. Typically, $B$ is initialized with zeros, while $A$ is initialized with a normal distribution. The output is then calculated as:
\begin{align*}
    \mathbf{h} = \mathcal{W}_{0}.\mathbf{X} + \Delta\mathcal{W}.\mathbf{X} \\
                = \mathcal{W}_{0}.\mathbf{X} + AB.\mathbf{X} 
\end{align*}
where $\mathbf{X}$ is the input to a layer or block. The dimensions of matrices $\mathbf{A}$ and $\mathbf{B}$ are determined by the LoRA 'r' and 'alpha' parameters. 

In our approach, we methodically identify the layers requiring updates during the fine-tuning stage through the LoRA process. To ensure stability and prevent training difficulties, we adopt techniques outlined by \citeauthor{lora} (\citeyear{lora}) and \citeauthor{tinygptv} (\citeyear{tinygptv}). Specifically, we fine-tune all attention layers and selectively update some fully connected layers. Additionally, we also fine-tune out projection layers to further enhance stability and robustness during the fine-tuning process, depicted in Figure \ref{fig:lora}. This systematic approach not only streamlines the fine-tuning process but also contributes to the overall stability and efficiency of our model.

\begin{table}
\centering
\renewcommand{\arraystretch}{1.5}
\begin{tabularx}{0.45\textwidth} { 
  >{\raggedright\arraybackslash}X 
  | >{\centering\arraybackslash}X 
  | >{\centering\arraybackslash}X }
 \textbf{Race} & \textbf{No. of Actions} & \textbf{No. of Units}\\ 
 \hline
 \hline
 Terran & 75 & 336 \\
 \hline
 Protoss & 61 & 246 \\
 \hline
 Zerg & 74 & 714 \\
\end{tabularx}
\caption{Build order actions choices by race.}
\label{table:bo}
\end{table}

While LoRA focuses on estimating the weights through smaller matrices, QLoRA leverages quantization techniques to reduce the memory footprint of the LLMs during fine-tuning while maintaining performance. Quantization involves reducing the precision of the model's parameters to lower bit widths, typically 8-bit or lower. In our case, we use 8-bit quantization during fine-tuning. By quantizing the model's parameters and applying layer-wise random adaptation, QLoRA enables fine-tuning of the entire LLM on a single GPU, making it suitable for deployment on devices with limited computational resources.

\renewcommand{\arraystretch}{1.6}
\begin{table}[ht]
\centering

\begin{tabularx}{0.45\textwidth} { 
  >{\raggedright\arraybackslash}p{0.06\textwidth} 
  | >{\centering\arraybackslash}p{0.08\textwidth}
  | >{\centering\arraybackslash}p{0.11\textwidth}
  | >{\centering\arraybackslash}p{0.1\textwidth}}
 
 \textbf{Games} & \textbf{GRU} & \textbf{Transformer} & \textbf{Ours}\\  
 \hline
 \multicolumn{4}{c}{\textbf{Mirror matchups}} \\
 \hline
 
 TvT & GS: 50.9 BO: 73.1 & GS: 52.23 BO: 74.38 & \textbf{GS: 54.88 BO: 76.82} \\
 \hline
 PvP & GS: 57.8 BO: 74.2 & GS: 58.42 BO: 74.6 & \textbf{GS: 61.08 BO: 78.49} \\
 \hline
 ZvZ & GS: 54.7 BO: 74.9 & GS: 56.01 BO: 74.6 & \textbf{GS: 58.27 BO: 77.07} \\
 \hline
 \multicolumn{4}{c}{\textbf{Non-mirror matchups}} \\
 \hline
PvT & GS: 57.0  BO: 69.6 & GS: 58.63 BO: 77.58 & \textbf{GS: 61.19 BO: 79.62} \\
\hline
PvT & GS: 56.9  BO: 74.2 & GS: 56.19 BO: 77.92 & \textbf{GS: 59.07 BO: 80.37} \\
\hline
TvZ & GS: 56.1  BO: 74.8 & GS: 55.79 BO: 75.22 & \textbf{GS: 60.46 BO: 78.74} \\
\end{tabularx}
\caption{Accuracy for global state (GS) and build order (BO) prediction. Our method outperforms previous work across all the match-ups.}
\label{table:result}
\end{table}

\section{Results}

\renewcommand{\arraystretch}{1.3}
\begin{table}[!ht]
\centering

\begin{tabularx}{0.45\textwidth} { 
  >{\raggedright\arraybackslash}p{0.1\textwidth} 
  | >{\centering\arraybackslash}p{0.11\textwidth}
  | >{\centering\arraybackslash}p{0.11\textwidth}}
 
 \textbf{Games} & \textbf{Fine-tuned Adapt} & \textbf{Zero-shot Transfer Learning} \\  
 \hline
 \hline
 
 TvT to TvZ & \textbf{GS: 54.88 BO: 76.82} & GS: 41.12 BO: 53.34\\ 
 \hline
 PvP to PvT & \textbf{GS: 61.08 BO: 78.49} & GS: 40.06 BO: 51.37\\
 \hline
 ZvZ to PvZ & \textbf{GS: 58.27 BO: 77.07} & GS: 39.41 BO: 51.74\\

\end{tabularx}
\caption{Accuracy for global state (GS) and build order (BO) prediction with zero-shot. Here the fine-tuned adapters are the same as in table \ref{table:result}.}
\label{table:result_tf}
\end{table}

\begin{table*} [!htp]
\centering
\renewcommand{\arraystretch}{1.5}
\begin{tabularx}{0.95\textwidth} { 
  >{\raggedright\arraybackslash}p{0.15\textwidth} 
  | >{\raggedright\arraybackslash}p{0.75\textwidth} }

    Ground Truth Actions & 'Research\_RavenCorvidReactor\_quick', 'Research\_AdvancedBallistics\_quick', 'Research\_RavenCorvidReactor\_quick', 'Research\_AdvancedBallistics\_quick' \\
    \hline
    
    Ground Truth Outcome & win \\
    \hline
    Generated Actions and Outcome &  Instruct: As an expert StarCraft II Terran player, playing against the Terran, predict the next 4 actions and also the result of the game, given the following resources:  \\ 
    & Game Stage: Mid, Army Count: low, Army Units/Buildings: 5 buildings \\
    & Minerals collected: low, Minerals used: low, Vespene gas collected: low, Vespene gas used: low \\
    & Food used: low, Food cap: low, Food for Army: low, Food for Workers: low \\
    & Idle Workers: low, Warp gates count: low, Larva count: low. \\
    & Output: \\                  
    & Action 1: Research\_RavenCorvidReactor\_quick \\
    & Action 2: Research\_AdvancedBallistics\_quick \\ 
    & Action 3: Research\_RavenCorvidReactor\_quick \\ 
    & \textcolor{red}{Action 4: Research\_RavenCorvidReactor\_quick} \\
    & Result: win
\end{tabularx}
\caption{A sample of ground truth actions and outcome along with model generated actions and outcome.}
\label{table:sample_res}
\end{table*}

For the initial set of experiments, we fine-tuned three distinct PEFT adapters: i) one for Terran vs. Terran matches; ii) one for Protoss vs. Protoss matches; and iii) one for Zerg vs. Zerg matches, following the first stage of our method. Each adapter was fine-tuned for an additional 2 to 3 epochs using a subset of their respective training replays as described in the training section. After fine-tuning, we evaluated each adapter on the test replays from the MSC dataset. Using the evaluation prompt,  we instructed the model to generate actions and predict the game outcome based on the provided information. The generated results were then compared to previous methods, with results summarized in Table \ref{table:result}. SC-Phi2 outperformed previous supervised approaches based on GRUs~\cite{msc} and transformers~\cite{junaid0} across all three match-ups. 

A sample of the generated actions and game outcome, along with ground truth actions and actual outcome is presented Table \ref{table:sample_res}. The sample shows both correctly generated actions and an incorrectly generated action (Action 4, marked in red). 

The next set of experiments explores the generalizability of the adapters. We took the adapters from the previous experiments and fine-tuned them for different match-ups. For instance, the adapter fine-tuned on the Terran vs. Terran match-up was tested using a zero-shot approach on Terran vs. Zerg match-up. The results are presented in Table \ref{table:result_tf}. These findings indicate that zero-shot transfer learning was not as effective as anticipated and that fine-tuned adapters provide a significant boost to performance.

\section{Conclusion}

In this work, we introduce SC-Phi2, a multimodal small language model that leverages both the Phi-2 and ViT models for SC2 macromanagement prediction tasks. Our approach employs dynamic prompts constructed from the game's global information, such as resources and food, along with textual descriptions of visual features extracted from the ViT model. These prompts are updated during fine-tuning to reflect the game's progress. Our method outperforms previous approaches in both global state prediction and build order prediction. In addition, we show that we can train our model on single GPU using LoRA and quantization approaches. 

Our research on fine-tuning small language models is a step towards reducing the carbon footprint of AI agents.
We concur with \citeauthor{gallotta2024largelanguagemodelsgames} (\citeyear{gallotta2024largelanguagemodelsgames}) that the most fertile areas for LLM research are likely to be in design and commentator systems rather than in surpassing the best AI players.
In future work, we plan to explore the usage of SC-Phi2 as a StarCraft commentator system that can comment on SC2 gameplay in real-time; prior work in this area~\cite{ranella2023towards} has demonstrated the utility of LLM commentary for League of Legends. 
 
\bibliography{references}

\begin{thebibliography}{27}
\providecommand{\natexlab}[1]{#1}

\bibitem[{Dettmers et~al.(2024)Dettmers, Pagnoni, Holtzman, and Zettlemoyer}]{qlora}
Dettmers, T.; Pagnoni, A.; Holtzman, A.; and Zettlemoyer, L. 2024.
\newblock QLORA: Efficient Finetuning of Quantized LLMs.
\newblock In \emph{Proceedings of the 37th International Conference on Neural Information Processing Systems}, NIPS '23. Red Hook, NY, USA: Curran Associates Inc.

\bibitem[{Dosovitskiy et~al.(2021)Dosovitskiy, Beyer, Kolesnikov, Weissenborn, Zhai, Unterthiner, Dehghani, Minderer, Heigold, Gelly, Uszkoreit, and Houlsby}]{vit}
Dosovitskiy, A.; Beyer, L.; Kolesnikov, A.; Weissenborn, D.; Zhai, X.; Unterthiner, T.; Dehghani, M.; Minderer, M.; Heigold, G.; Gelly, S.; Uszkoreit, J.; and Houlsby, N. 2021.
\newblock An Image is Worth 16x16 Words: Transformers for Image Recognition at Scale.
\newblock \emph{ICLR}.

\bibitem[{Gallotta et~al.(2024)Gallotta, Todd, Zammit, Earle, Liapis, Togelius, and Yannakakis}]{gallotta2024largelanguagemodelsgames}
Gallotta, R.; Todd, G.; Zammit, M.; Earle, S.; Liapis, A.; Togelius, J.; and Yannakakis, G.~N. 2024.
\newblock Large Language Models and Games: A Survey and Roadmap.
\newblock arXiv:2402.18659.

\bibitem[{Gunasekar et~al.(2023)Gunasekar, Zhang, Aneja, Mendes, Giorno, Gopi, Javaheripi, Kauffmann, de~Rosa, Saarikivi, Salim, Shah, Behl, Wang, Bubeck, Eldan, Kalai, Lee, and Li}]{phi2}
Gunasekar, S.; Zhang, Y.; Aneja, J.; Mendes, C. C.~T.; Giorno, A.~D.; Gopi, S.; Javaheripi, M.; Kauffmann, P.; de~Rosa, G.; Saarikivi, O.; Salim, A.; Shah, S.; Behl, H.~S.; Wang, X.; Bubeck, S.; Eldan, R.; Kalai, A.~T.; Lee, Y.~T.; and Li, Y. 2023.
\newblock Textbooks Are All You Need.
\newblock arXiv:2306.11644.

\bibitem[{Hu et~al.(2022)Hu, Shen, Wallis, Allen-Zhu, Li, Wang, Wang, and Chen}]{lora}
Hu, E.~J.; Shen, Y.; Wallis, P.; Allen-Zhu, Z.; Li, Y.; Wang, S.; Wang, L.; and Chen, W. 2022.
\newblock Lo{RA}: Low-Rank Adaptation of Large Language Models.
\newblock \emph{ICLR 22}.

\bibitem[{Hu et~al.(2024)Hu, Huang, Ilhan, Tekin, Liu, Kompella, and Liu}]{survey_llm}
Hu, S.; Huang, T.; Ilhan, F.; Tekin, S.; Liu, G.; Kompella, R.; and Liu, L. 2024.
\newblock A Survey on Large Language Model-Based Game Agents.
\newblock arXiv:2404.02039.

\bibitem[{Huang et~al.(2023)Huang, Wu, Yu, Fan, Fu, Fu, and Yang}]{roa_star}
Huang, R.; Wu, X.; Yu, H.; Fan, Z.; Fu, H.; Fu, Q.; and Yang, W. 2023.
\newblock A Robust and Opponent-Aware League Training Method for StarCraft II.
\newblock In Oh, A.; Naumann, T.; Globerson, A.; Saenko, K.; Hardt, M.; and Levine, S., eds., \emph{Advances in Neural Information Processing Systems}, volume~36, 47554--47574. Curran Associates, Inc.

\bibitem[{Kambhampati(2024)}]{Kambhampati_2024}
Kambhampati, S. 2024.
\newblock Can large language models reason and plan?
\newblock \emph{Annals of the New York Academy of Sciences}, 1534(1): 15–18.

\bibitem[{Khan, Hassan, and Sukthankar(2021)}]{junaid0}
Khan, M.~J.; Hassan, S.; and Sukthankar, G. 2021.
\newblock Leveraging Transformers for StarCraft Macromanagement Prediction.
\newblock In \emph{IEEE International Conference on Machine Learning and Applications (ICMLA)}, 1229--1234.

\bibitem[{Li et~al.(2023)Li, Li, Savarese, and Hoi}]{blip2}
Li, J.; Li, D.; Savarese, S.; and Hoi, S. 2023.
\newblock {BLIP-2}: bootstrapping language-image pre-training with frozen image encoders and large language models.
\newblock \emph{International Conference on Machine Learning}.

\bibitem[{Liquipedia(2024)}]{liquipedia}
Liquipedia. 2024.
\newblock {Liquipedia}.
\newblock \url{https://liquipedia.net/starcraft/Main_Page}.
\newblock {Accessed: Feb 16, 2024}.

\bibitem[{Ma et~al.(2024)Ma, Mi, Zeng, Yan, Wu, Lin, Zhang, and Wang}]{textsc2}
Ma, W.; Mi, Q.; Zeng, Y.; Yan, X.; Wu, Y.; Lin, R.; Zhang, H.; and Wang, J. 2024.
\newblock Large Language Models Play StarCraft II: Benchmarks and A Chain of Summarization Approach.
\newblock arXiv:2312.11865.

\bibitem[{Radford et~al.(2021)Radford, Kim, Hallacy, Ramesh, Goh, Agarwal, Sastry, Askell, Mishkin, Clark, Krueger, and Sutskever}]{clip}
Radford, A.; Kim, J.~W.; Hallacy, C.; Ramesh, A.; Goh, G.; Agarwal, S.; Sastry, G.; Askell, A.; Mishkin, P.; Clark, J.; Krueger, G.; and Sutskever, I. 2021.
\newblock Learning Transferable Visual Models From Natural Language Supervision.
\newblock arXiv:2103.00020.

\bibitem[{Ranella and Eger(2023)}]{ranella2023towards}
Ranella, N.; and Eger, M. 2023.
\newblock Towards Automated Video Game Commentary Using Generative AI.
\newblock In \emph{EXAG@ AIIDE}.

\bibitem[{Shao et~al.(2024)Shao, Jiang, Zuo, and Liu}]{swarmbrain}
Shao, X.; Jiang, W.; Zuo, F.; and Liu, M. 2024.
\newblock SwarmBrain: Embodied agent for real-time strategy game StarCraft II via large language models.
\newblock arXiv:2401.17749.

\bibitem[{{StarCraft - Wikipedia}(2024)}]{scwikipedia}
{StarCraft - Wikipedia}. 2024.
\newblock {StarCraft - Wikipedia}.
\newblock \url{https://en.wikipedia.org/wiki/StarCraft}.
\newblock {Accessed: Feb 20, 2024}.

\bibitem[{Vinyals et~al.(2019)Vinyals, Babuschkin, Czarnecki, Mathieu, Dudzik, Chung, Choi, Powell, Ewalds, Georgiev, Oh, Horgan, Kroiss, Danihelka, Huang, Sifre, Cai, Agapiou, Jaderberg, Vezhnevets, Leblond, Pohlen, Dalibard, Budden, Sulsky, Molloy, Paine, Gulcehre, Wang, Pfaff, Wu, Ring, Yogatama, Wünsch, McKinney, Smith, Schaul, Lillicrap, Kavukcuoglu, Hassabis, Apps, and Silver}]{alpha}
Vinyals, O.; Babuschkin, I.; Czarnecki, M.~W.; Mathieu, M.; Dudzik, A.; Chung, J.; Choi, H.~D.; Powell, R.; Ewalds, T.; Georgiev, P.; Oh, J.; Horgan, D.; Kroiss, M.; Danihelka, I.; Huang, A.; Sifre, L.; Cai, T.; Agapiou, P.~J.; Jaderberg, M.; Vezhnevets, S.~A.; Leblond, R.; Pohlen, T.; Dalibard, V.; Budden, D.; Sulsky, Y.; Molloy, J.; Paine, L.~T.; Gulcehre, C.; Wang, Z.; Pfaff, T.; Wu, Y.; Ring, R.; Yogatama, D.; Wünsch, D.; McKinney, K.; Smith, O.; Schaul, T.; Lillicrap, T.; Kavukcuoglu, K.; Hassabis, D.; Apps, C.; and Silver, D. 2019.
\newblock Grandmaster level in {StarCraft II} using multi-agent reinforcement learning.
\newblock \emph{Nature}, 1--5.

\bibitem[{Vinyals et~al.(2017)Vinyals, Ewalds, Bartunov, Georgiev, Vezhnevets, Yeo, Makhzani, K{\"{u}}ttler, Agapiou, Schrittwieser, Quan, Gaffney, Petersen, Simonyan, Schaul, van Hasselt, Silver, Lillicrap, Calderone, Keet, Brunasso, Lawrence, Ekermo, Repp, and Tsing}]{sc2le}
Vinyals, O.; Ewalds, T.; Bartunov, S.; Georgiev, P.; Vezhnevets, A.~S.; Yeo, M.; Makhzani, A.; K{\"{u}}ttler, H.; Agapiou, J.~P.; Schrittwieser, J.; Quan, J.; Gaffney, S.; Petersen, S.; Simonyan, K.; Schaul, T.; van Hasselt, H.; Silver, D.; Lillicrap, T.~P.; Calderone, K.; Keet, P.; Brunasso, A.; Lawrence, D.; Ekermo, A.; Repp, J.; and Tsing, R. 2017.
\newblock StarCraft {II:} {A} New Challenge for Reinforcement Learning.
\newblock \emph{CoRR}, abs/1708.04782.

\bibitem[{Wang et~al.(2023)Wang, Xie, Jiang, Mandlekar, Xiao, Zhu, Fan, and Anandkumar}]{voyager}
Wang, G.; Xie, Y.; Jiang, Y.; Mandlekar, A.; Xiao, C.; Zhu, Y.; Fan, L.; and Anandkumar, A. 2023.
\newblock Voyager: An Open-Ended Embodied Agent with Large Language Models.
\newblock \emph{Trans. Mach. Learn. Res.}, 2024.

\bibitem[{Wei et~al.(2022)Wei, Wang, Schuurmans, Bosma, Ichter, Xia, Chi, Le, and Zhou}]{cot}
Wei, J.; Wang, X.; Schuurmans, D.; Bosma, M.; Ichter, B.; Xia, F.; Chi, E.; Le, Q.~V.; and Zhou, D. 2022.
\newblock Chain-of-Thought Prompting Elicits Reasoning in Large Language Models.
\newblock In Koyejo, S.; Mohamed, S.; Agarwal, A.; Belgrave, D.; Cho, K.; and Oh, A., eds., \emph{Advances in Neural Information Processing Systems}, volume~35, 24824--24837. Curran Associates, Inc.

\bibitem[{Wiki(2024)}]{scwiki}
Wiki, S. 2024.
\newblock {StarCraft Wiki}.
\newblock \url{https://starcraft.fandom.com/wiki/StarCraft_Wiki#}.
\newblock {Accessed: Feb 18, 2024}.

\bibitem[{Wolf et~al.(2020)Wolf, Debut, Sanh, Chaumond, Delangue, Moi, Cistac, Rault, Louf, Funtowicz, Davison, Shleifer, von Platen, Ma, Jernite, Plu, Xu, Scao, Gugger, Drame, Lhoest, and Rush}]{huggingfaces}
Wolf, T.; Debut, L.; Sanh, V.; Chaumond, J.; Delangue, C.; Moi, A.; Cistac, P.; Rault, T.; Louf, R.; Funtowicz, M.; Davison, J.; Shleifer, S.; von Platen, P.; Ma, C.; Jernite, Y.; Plu, J.; Xu, C.; Scao, T.~L.; Gugger, S.; Drame, M.; Lhoest, Q.; and Rush, A.~M. 2020.
\newblock HuggingFace's Transformers: State-of-the-art Natural Language Processing.
\newblock arXiv:1910.03771.

\bibitem[{Wu, Zhang, and Huang(2017)}]{msc}
Wu, H.; Zhang, J.; and Huang, K. 2017.
\newblock MSC: A Dataset for Macro-Management in StarCraft II.
\newblock \emph{arXiv preprint arXiv:1710.03131}.

\bibitem[{Yuan et~al.(2023)Yuan, Zhang, Wang, Xie, Cai, Dong, and Lu}]{open_world_long_horizon}
Yuan, H.; Zhang, C.; Wang, H.; Xie, F.; Cai, P.; Dong, H.; and Lu, Z. 2023.
\newblock Skill Reinforcement Learning and Planning for Open-World Long-Horizon Tasks.
\newblock In \emph{NeurIPS 2023 Foundation Models for Decision Making Workshop}.

\bibitem[{Yuan, Li, and Sun(2023)}]{tinygptv}
Yuan, Z.; Li, Z.; and Sun, L. 2023.
\newblock TinyGPT-V: Efficient Multimodal Large Language Model via Small Backbones.
\newblock arXiv:2312.16862.

\bibitem[{Zhou et~al.(2024)Zhou, Qin, Yin, Huang, Zhang, Sheng, Qiao, and Shao}]{minedreamer}
Zhou, E.; Qin, Y.; Yin, Z.; Huang, Y.; Zhang, R.; Sheng, L.; Qiao, Y.; and Shao, J. 2024.
\newblock MineDreamer: Learning to Follow Instructions via Chain-of-Imagination for Simulated-World Control.
\newblock arXiv:2403.12037.

\bibitem[{Zhu et~al.(2023)Zhu, Chen, Tian, Tao, Su, Yang, Huang, Li, Lu, Wang, Qiao, Zhang, and Dai}]{mc_ghost}
Zhu, X.; Chen, Y.; Tian, H.; Tao, C.; Su, W.; Yang, C.; Huang, G.; Li, B.; Lu, L.; Wang, X.; Qiao, Y.; Zhang, Z.; and Dai, J. 2023.
\newblock Ghost in the Minecraft: Generally Capable Agents for Open-World Environments via Large Language Models with Text-based Knowledge and Memory.
\newblock arXiv:2305.17144.

\end{thebibliography}
\clearpage
\onecolumn
\section{Supplementary Material}

For both stage 1 and stage 2 fine-tuning experiments, we utilize a system equipped with an NVIDIA RTX 3090 GPU with 24GB of VRAM, an Intel Core i7-11700KF CPU with 16 cores, and 64GB of system RAM.

\subsection{Stage-1 Self-Supervised Fine-Tuning}

The following table presents the fine-tuning details for Stage 1, including various hyperparameters and the model's performance under each parameter configuration. Across all configurations, we maintain a batch size of 1, utilize a cosine learning rate scheduler, implement 8 gradient accumulation steps, set a token length of 820, and employ Flash attention.

Extending the fine-tuning duration invariably enhances performance, albeit at the cost of increased training time. Additionally, incorporating warmup steps has a positive impact on the model's fine-tuning performance, as evidenced by the highlighted row in the table. Moreover, the non-paged version of the AdamW optimizer outperformed its paged counterpart, even when operating in 8-bit mode. 

\begingroup
\renewcommand{\arraystretch}{1.6}
\begin{table*}[!h]
\centering
\begin{tabularx}{\textwidth} { 
   >{\centering\arraybackslash}p{0.055\textwidth} |
  >{\centering\arraybackslash }p{0.065\textwidth} 
  | >{\centering\arraybackslash}p{0.08\textwidth} 
  | >{\centering\arraybackslash}p{0.08\textwidth}
  | >{\centering\arraybackslash}p{0.07\textwidth}
  | >{\centering\arraybackslash}p{0.05\textwidth} 
  | >{\centering\arraybackslash}p{0.05\textwidth} 
  | >{\centering\arraybackslash}p{0.09\textwidth}
  | >{\centering\arraybackslash}p{0.18\textwidth}}
 \textbf{LoRA r} & \textbf{LoRA alpha} & \textbf{Training Time} & \textbf{Training Loss} & \textbf{Epochs} & \textbf{4 Bit} & \textbf{8 Bit} & \textbf{Warmup Steps} & \textbf{Optimizer}
    \\  
    \hline
    \hline
    32 & 64 & 9625 & 1.58 & 40 & Yes & No & 0 & Paged AdamW 8bit 
    \\
    64 & 128 & 10587 & 2.0427 & 20 & Yes & No & 20 & Paged AdamW 8bit
    \\
    64 & 128 & 20946 & 1.57 & 40 & Yes & No & 20 & Paged AdamW 8bit  
    \\
    \hline
    64 & 128 & 22157 & 1.56 & 40 & No & Yes & 20 & Paged AdamW 8bit 
    \\
    64 & 128 & 43329 & 1.06 & 80 & No & Yes & 30 & AdamW 8bit 
    \\
    64 & 128 & 54705 & 0.8854 & 100 & No & Yes & 30 & AdamW 8bit 
    \\
    \hline
    64 & 128 & 54277 & 0.9866 & 100 & No & Yes & 30 & Paged AdamW 32bit 
    \\
     64 & 128 & 78544 & 0.6355 & 140 & No & Yes & 50 & AdamW 8bit 
    \\
    \textbf{64} & \textbf{128} & \textbf{87153} & \textbf{0.5564} & \textbf{160} & \textbf{No} & \textbf{Yes} & \textbf{80} & \textbf{AdamW 8bit} 
\end{tabularx}
\caption{Comparison of Stage-1 fine-tuning across different hyperparameter configurations (training times are given in seconds).}
\label{table:app_stg1}
\end{table*}
\endgroup

Optimizing the values of the LoRA hyperparameters 'r' and 'alpha' significantly influences the fine-tuning process of large language models (LLMs). Larger values for both hyperparameters can enhance fine-tuning performance, albeit at the expense of increased memory requirements and the number of trainable parameters. Detailed statistics on these aspects are provided in Table \ref{table:t_params}.

\begingroup
\renewcommand{\arraystretch}{1.5}
\begin{table*}[h]
\centering
\begin{tabularx}{\textwidth} { 
   >{\centering\arraybackslash}p{0.15\textwidth} |
  >{\centering\arraybackslash }p{0.1\textwidth} 
  | >{\centering\arraybackslash}p{0.1\textwidth} 
  | >{\centering\arraybackslash}p{0.2\textwidth}
  | >{\centering\arraybackslash}p{0.1\textwidth}
  | >{\centering\arraybackslash}p{0.15\textwidth}}
 \textbf{Base Model Params} & \textbf{LoRA r} & \textbf{LoRA alpha} & \textbf{Trainable Params} & \textbf{Token Length} & \textbf{GPU Memory}
    \\  
    \hline
    \hline
    2,889,784,320 & 32 & 64 & 1.31\% | 36,700,160 & 820 & 17.1GB \\
    2,889,784,320 & 64 & 128 & 2.57\% | 74,400,320 & 820 & 17.5GB \\
    2,889,784,320 & 96 & 192 & 3.81\% | 110,100,480 & 820 & 17.9GB 
\end{tabularx}
\caption{Comparison of various values of 'r' and 'alpha' and their impact on trainable parameters and GPU memory.}
\label{table:t_params}
\end{table*}
\endgroup
\clearpage
\subsection{Stage-2 Self-Supervised Fine-Tuning}
Following the best parameters listed in the Stage-1 fine-tuning, we proceed with the Stage-2 fine-tuning. Again, we set LoRA 'alpha' to 192, and 'r' to 96. However, we change the batch size to 4, and token length to 288, but keep gradient accumulation steps same as well as the optimizer. 
Table~\ref{table:s2_params} lists the stage 2 fine-tuning configuration. 

\begingroup
\renewcommand{\arraystretch}{1.5}
\begin{table*}[htp]
\centering
\begin{tabularx}{\textwidth} { 
   >{\centering\arraybackslash}p{0.13\textwidth} |
  >{\centering\arraybackslash }p{0.1\textwidth} 
  | >{\centering\arraybackslash}p{0.1\textwidth} 
  | >{\centering\arraybackslash}p{0.17\textwidth}
  | >{\centering\arraybackslash}p{0.1\textwidth}
  | >{\centering\arraybackslash}p{0.1\textwidth}
  | >{\centering\arraybackslash}p{0.13\textwidth}}
 \textbf{Base model Params} & \textbf{LoRA r} & \textbf{LoRA alpha} & \textbf{Trainable Params} & \textbf{Token Length} & \textbf{Batch Size} & \textbf{GPU Memory}
    \\  
    \hline
    \hline
    2,889,784,320 & 96 & 192 & 3.81\% | 110,100,480 & 288 & 4 & 23.44GB 
\end{tabularx}
\caption{Stage 2 training configuration, keeping the best configuration of stage 1, except batch size and token length.}
\label{table:s2_params}
\end{table*}
\endgroup



\subsection{SC2 Text Dataset}

Our dataset consists of 1,500 instances and is available upon request. Designed in a question-answer format, it is well-suited for fine-tuning language models, particularly with the instructional prompt format for Phi-2. Additionally, it is compatible with LLaMA-style instructional fine-tuning. Table \ref{table:dataset} provides some examples from the dataset.

\begingroup
\renewcommand{\arraystretch}{1.5}
\begin{table*}[htp]
\centering
\begin{tabularx}{\textwidth} { 
   >{\raggedright}p{0.25\textwidth} |
  >{\raggedright\arraybackslash }p{0.7\textwidth}}
    \centering{\textbf{Question}} & \textbf{Answer}
    \\  
    \hline
    \hline
    What is the role of Extractor in Zerg? & The Extractor is the Zerg building from which Drones collect Vespene Gas. The Extractor is the Zerg equivalent to the Terran Refinery and the Protoss Assimilator. Besides being the fundamental Zerg building for collecting the Vespene Gas resource, the Extractor is sometimes used early game to bypass the 14 food limit by using the Extractor Trick. Players can build one or two Extractors at their main base, start the production of one or two Drones and then cancel the Extractor(s) as a way of avoiding the delay of a morphing Overlord when supply capped. This temporary decrease of supply is only possible for Zerg due to the fact that a morphing Extractor, along with all other Zerg buildings, requires the sacrifice of one Drone. Note that performing this supply trick will only refund 19 of the 25 minerals (75\% rounded) used to morph the Extractor. Scouting or expanding drones that are under siege can be morphed into extractors on the fly to prevent the drone from being killed. They can later be canceled to continue scouting or expanding once the threats have been nullified. \\

    What is the role of Terran Medivac vs. Protoss? & Medivacs are brought with a Terran bio army to provide healing support. They are also frequently used to drop units in the Protoss base and snipe important infrastructure (Mining Probes, Pylons, Nexus, tech structures). They are also used in TvP for Hellion/Hellbat drops. \\ 

    What are the Training actions in Protoss? & 
    
        "Train\_Adept\_quick", "Train\_Carrier\_quick", "Train\_Colossus\_quick", "Train\_DarkTemplar\_quick", "Train\_Disruptor\_quick", "Train\_HighTemplar\_quick", "Train\_Immortal\_quick", "Train\_MothershipCore\_quick", "Train\_Observer\_quick", "Train\_Oracle\_quick", "Train\_Phoenix\_quick", "Train\_Probe\_quick", "Train\_Sentry\_quick", "Train\_Stalker\_quick", "Train\_Tempest\_quick", "Train\_VoidRay\_quick", "Train\_WarpPrism\_quick", "Train\_Zealot\_quick"
\end{tabularx}
\caption{Examples from our SC2 Text Dataset}
\label{table:dataset}
\end{table*}
\endgroup

\end{document}